\documentclass[lnbip]{svmultln}
\usepackage{array}
\usepackage{graphicx}
\usepackage{amsfonts} 
\usepackage{makeidx}  
\begin{document}
\mainmatter              
\title{Targeted change detection in remote sensing images}
\titlerunning{Targeted change detection}  
%
\author{Vladimir Ignatiev\inst{1} \and Alexey Trekin\inst{1} \and Viktor Lobachev \inst{1} \and Georgy Potapov\inst{1} \and Evgeny Burnaev\inst{1}}
\authorrunning{Vladimir Ignatiev et al.}   
%
\tocauthor{Vladimir Ignatiev, Alexey Trekin, Viktor Lobachev, Georgy Potapov, Evgeny Burnaev}
\institute{Skolkovo institute of Science and technology, Moscow, Nobel st. 1, Russia,\\
\email{aeronetlab@skoltech.ru},\\ WWW home page:
\texttt{http://crei.skoltech.ru/cdise/aeronet-lab/}}

\maketitle              

\begin{abstract}        
Recent developments in the remote sensing systems and image processing made it possible to propose a new method of the object classification and detection of the specific changes in the series of satellite Earth images (so called targeted change detection). 
In this paper we propose a formal problem statement that allows to use effectively the deep learning approach to analyze time-dependent series of remote sensing images.
We also introduce a new framework for the development of deep learning models for targeted change detection and demonstrate some cases of business applications it can be used for.
\end{abstract}
\section{Problem overview}
Earth observation systems have been developing during the last decades in terms of quality of the imagery and frequency of the imaging \cite{toth}. However, until now the daily basis imagery wasn’t available for any location or area of interest. Each satellites and UAV systems have their own limitations - viz. number of satellites, weather and daylight for optical systems, vegetation for SAR systems, not to say the commercial ones. That lead the remote sensing industry to the situation where the amount of the data and the proclaimed frequency of the repeating imaging could have given us full daily coverage of the earth surface by the high-resolution imagery \cite{planet}, but in the reality it turns out to be composed of the diverse, inconsistent and often incomplete data.
And yet that business and technology trend raise the issues how all this daily imagery can be targeted to the specific needs and used in a various types of applications. The technical issues that can be raised following these business needs are:
\begin{itemize}
\item data fusion of different types of remote sensing data coming from different sensors (optical, SAR, different resolution etc.) - to provide the maximum capacity of operational monitoring in any time or weather conditions \cite{dong09, zhang10} 
\item detection of changes in the time series of imagery and in a very short time spans to provide near real time valuable information related to the objects of interest. As the traditional approaches fail to make this massive dataflow a reliable source of the useful answers we should propose the new machine learning approaches that can handle it and provide the automated analysis that can compete to the human one by accuracy
\end{itemize}

The business needs lead to the concept of the targeted change detection (TCD) that is consolidated detection and classification of the changes in order to obtain only the interesting for the problem (target) changes rather than all the changes in general \cite{fernandez, ye}.

Our motivation in this paper is to build a general description of change detection (CD) problem taking into account characteristics of input data - series of remote sensing (RS) images - and information about ``targets'' obtained from classification of specific business applications of the respective territories such as ``transportation \& logistic'', ``residential areas'', ``building \& construction'' etc. Having formalized problem statement it is possible to develop overall framework that will reduce costs and simplify stages of collecting data, model construction and integration in workflow of business solutions.
\subsection{Business applications of CD}

Commercial services producing analytics based on series of multitemporal remote sensing images begin occupying significant share in such market areas as smart farming \cite{gamaya, digitalglobe, e-farming}, monitoring of security zones of linear objects that have enormous transport network \cite{I-Cubed, SIC, orbilateye}, monitoring of protected areas \cite{sovzond, unmanned_systems, zala}, assessment of damage caused by natural and man-made disasters \cite{ , }, analytical solutions for consulting agencies in geo-marketing \cite{rilos,GIM}.

Our survey of the commercial solutions based on analysis of RS images reveals the following problems:
\begin{itemize}
\item only a number of applications have algorithmic solutions;
\item most of solutions do not have enough accuracy; 
\item the plenty of problems do not have any reliable solutions;
\item companies are in search of new frameworks that will anticipate their risks or will make precise and timely estimation of losses. 
\end{itemize}

It is obvious that each of use cases has its own demands and requirements to the changes that should be detected. For example, if one is going to monitor security zones of pipelines, he should find out the list of targets, changes of which will be detected. Each target means a specific type of objects that has the characteristic linear size and this value defines the spatial resolution of RS images to be used. Most of real business applications requires to process with RS images of high and very high spatial resolution from 5 cm/pixel to 10 m/pixel. 

\subsection{RS datasets available for CD algorithms}
The idea of making general framework providing development of robust algorithms for detecting changes for the specific applied problem requires building representative datasets. 
There are scarce datasets for change detection on RS images nowadays \cite{AICD, SARcdds, GEEds, cdnet}. Only few of existing datasets can be implemented within the framework of targeted change detection \cite{AICD,cdnet}. Most of them have not enough coverage or time spread to learn and test algorithms using modern techniques of machine learning and deep learning.
So the one of the challenges of our work is to collect and build a number of datasets dedicated to the task of targeted CD. 

\subsection{Current methods and algorithms}
\label{methods}
Solutions proposed in the latest papers could be divided into several groups according to the used models and algorithms, to market domains and to RS images in use. 
Traditional methods of change detection by series of satellite images are typically limited in several aspects:
\begin{itemize}
\item using just two images for justification with ground truth;
\item do not take into account usability of changes in specific use case;
\item only empirical approaches for selection of the appropriate RS images for real applications;
\item no automated end-to-end solutions for business cases. 
\end{itemize}

One of the first applied research in this area was conducted by W.Roper and S.Dutta \cite{Roper}. They made an attempt to understand which kind of sensor system aerial or satellite is better in a sense of probability of detection unauthorized intrusion onto a pipeline right-of-way. The main conclusion of this work is that one should use aerial imagery when the imaging frequency is lower than once per month, but starting from once per day imaging frequency - probability of change detection with satellite imagery is higher. There were used standard algorithms of classification built-in such software packages as ENVI \cite{ENVI}.

The future works were focused on improving change detection algorithms themselves, but the idea of targeted CD were pushed into the background. So, the best classical approaches are methods based on anomaly detection framework on time series of multispectral low-resolution satellite images and so called spectral indices \cite{Cai2015}, methods based on Markov Random Fields and global optimization on graphs \cite{Vakalopoulou2015, Yu2017,Gu2017}, approaches using object-based segmentation with post-classification of changes \cite{Jianya2008,Huang2015,Vittek2014} and methods based on Multivariate Alteration Detection \cite{Jabari2016,Wang2017}. All the mentioned approaches shows quiet good results in quality, but either use multi framed series of low resolution satellite images or use high resolution images, but the same two pairs of images for both training and testing.     

Latest research achievements in computer vision made it possible to use deep neural networks for accurate semantic segmentation objects on daily images. The next step was to transfer this technology on remote sensing data. Actually, it’s ongoing research. This results are the basement for development new algorithms of CD. It can be emphasised the most successful studies in this direction. A.M. Amin et al. \cite{Amin2016} proposed a method for binary classification of changes based on CNN with input two high resolution satellite images (0.65 - 2.62 m/pixel) gives the accuracy by Kappa coefficient - 0.876. The second approach by Y. Chu et al. \cite{Chu2016}, based on clustering analysis of difference of two images (0.65 - 2.62 m/pixel) uses deep belief network model with restricted Boltzmann machine architecture shows the accuracy by Kappa coefficient from 0.905 to 0.987. The main advantage of this method that it allows to avoid radiometric correction of images, they outperform the standard approaches and could be adopted for targeted change detection in different use cases. Deep learning techniques need a lot of computational resources for training process and could be unstable for analysis of new images, but the last problem may be eliminated having a representative dataset for model training. 

Further researches described in \cite{Wang2017, Gu2017, sakurada, mcdermid, huang} give promising results, but most of them share the same drawback: low amount of data used for the method validation. The of creation a large and reliable dataset is one of the keys to the development of robust and reliable algorithms for the targeted change detection.



One of the important issues is that many real problems do not need any changes
to be detected, but only the interesting, problem-specific ones. This implies a
challenging task to develop a method that could extract only the specified classes of changes.
The deep neural networks appear to be the most promising approach to the
problem, but the current solutions are highly limited by the amount
of the data available for the training of the models. 

\section{Remote sensing data for change detection}
In order to understand the capability of the modern space imagery we need to
examine its characteristics. As stated in the previous section, according to the
target sizes we need the images in high (1-10 meters) and very high (less than 1 meter) spatial resolution. A typical high resolution satellite in 2018 has 4 or more spectral bands and a panchromatic image of higher spatial resolution (see table \ref{table_sat}). A completely new approach to the data acquisition is demonstrated by the PlanetScope satellite group, that consists of a large group of very small satellites (3-unit CubeSat), that can acquire far more imagery than traditional satellites of comparable mass and cost, however at the cost of less data quality: they have less spectral bands and the noise level is higher \cite{santilli}.

\begin{table}[htp]
\begin{center}
  \begin{tabular}{ | c | c| c | c| }
    \hline
     & SPOT-6 & PlanetScope & Worldview-3 \\ \hline
    Visible channels & 4 & 3 & 6 \\ \hline
    Near infrared channels & 1 & 0/1 & 2 \\ \hline
    SW infrared channels &0 & 0 & 8 \\ \hline
    Multispectral resolution, m & 6.0 & 3.0 / 3.7 & 1.24 / 3.72 \\ \hline
    Panchromatic resolution, m &1.5 & - & 0.31 \\ \hline
    Daily coverage, millions sq.km/ day & 3 & 150 & 0.68 \\
    \hline
  \end{tabular}
  \label{table_sat}
\end{center}\caption{Overview of the technical specifications of the modern RS satellites}
\end{table}


Let us call a set of images taken by one space or aerial instrument at one flight and covering the same territory a ``multi-layer image''. We need this term do differentiate from the multispectral image where each layer represents a spectral band and the panchromatic layer is not included into the multispectral image. In our case we add it as well as any other information represented in a raster form as a layer to our image representation, as we want to collect all the information within a single instance of a multi-layer image.

$I = {I_1, ... , I_C}$ is a multi-layer image with $C$ bands (channels). Every band $I_i$ is a single 2-dimensional matrix with the size $(X_i, Y_i)$. Different bands can have different spatial resolution so, under the condition that all the bands must cover the same territory,  the sizes of the channels are different, but proportional by an integer ratio. If we order the bands by the decrease of the size, their dimensions will be $X_i / X{i+1} = Y_i / Y_{i+1} \in \mathbb{N}$

For the task of change detection the images are organized in time series: $S = {I^1, ... I^T}$ , where each image $I^t, t = 1...T$ covers the same area, but different images can be obtained by different instruments and have different characteristics. We intentionally put this restriction on the area, because we need not to describe all the data available, but to construct the datasets that can be used by the data analysis methods within the proposed framework. If the given data are not fully spatially consistent, we can crop the extents and, if needed, divide the time series into several separate ones of different spatial and temporal extent.

\section{Proposed framework}

This section discusses details of proposed framework and shows the example of methods for targeted change detection in classical situation when two input satellite images are used for classification of one type of changes. 

\subsection{Problem statement}

Given the series of images $S = {I^1, ... I^T}$ where each image $I^t$ is a multi-layer image taken at the moment of time $t$, a parameterized description of changes occurred in the series can be introduced as $C=F(S)$. The exact definition of the operator $F$ is done at the post-processing stage and is beyond the scope of this paper. It is assumed that the calculation of the vector $C$ will use only the set of objects of pair-wise changes $\{r_{ij}\}$, so the problem goal can be reduced to the search of $r_{ij}$ - the change between the images $I^i$ and $I^j$.
The pair change object definition includes:
\begin{itemize}
\item Category of detected changes;
\item The set of polygonal domains (in physical coordinates) where the changes are detected.
\end{itemize}
Evaluation of the detected change is done with one of the measures:
\begin{itemize}
\item Intersection over Union;
\item The pair Recall-Precision;
\item F1-measure.
\end{itemize}
All three measures assume availability of the ground truth changes for the investigated pair of images. Each measure can be calculated either per unit area or per object; the latter is specified heuristically.

\subsection{Description of the framework for TCD applications}

Our framework is restricted neither by the type of remote sensing data, nor methods, algorithms, but it does take into account information about the ``targets'' of the applied area. 

The question of determining the context of using the notion ``change'' is removed by establishing a one-to-one correspondence of the model or algorithm for detecting changes and the business case to be solved.
Robustness of the created algorithms is guaranteed by creating reference sets of markup of changes for a series of different images and applying modern methods of in-depth training on data generated from reference sets.

\begin {figure}
\centering
\includegraphics[width=\textwidth]{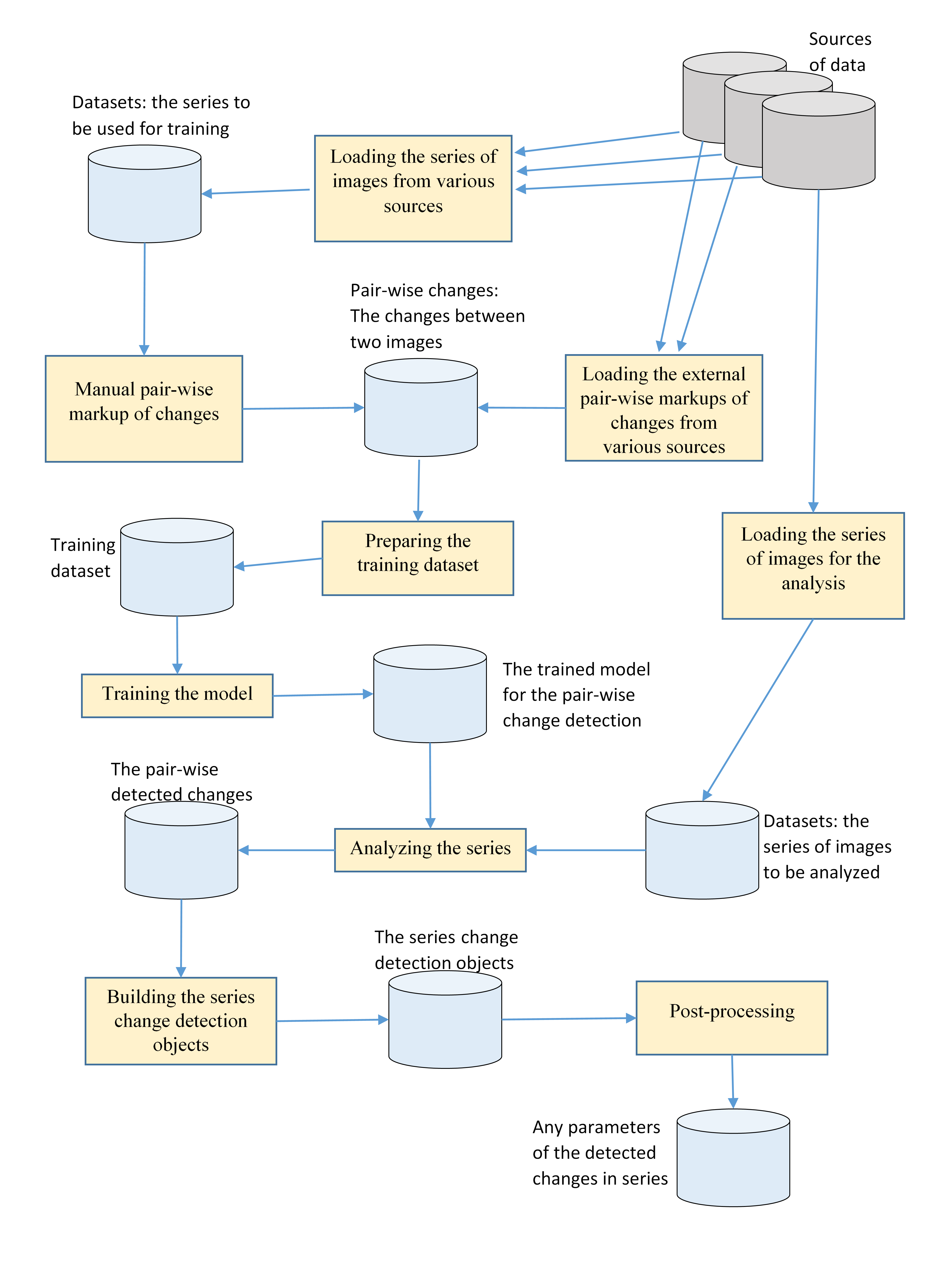}
\caption{Scheme of the proposed framework}
\label{framework}
\end{figure}

As one can see on the scheme depicted on the figure \ref{framework} there are three stages in proposed framework: preparing dataset, training a model and analyzing series of images for pair-wise targeted change detection. Dataset for specific business application should be constructed in accordance with the formal mathematical problem statement described above. It means that one should load series of images concerning the size of targets, the characteristic time period of occurrence of changes, type of images and so on. This data are combined with external mark-ups of changes of objects (targets) under consumption. The model for pair-wise targeted change detection is chosen from the set of neural networks architectures prepared in advance and is trained and validated on the dataset built of the first stage. The process of analyzing new series of images starts from building map of changes for each pair of images inside the series. After that there is a post-processing and verification procedures that resulted in recommendation, forecasts or alerting reports with detected changes, that can be visualized in interactive mode.

\subsection{Method of TCD based on deep learning techniques}

Our method is fundamentally different from all the methods listed in section \ref{methods} because it takes into account information about targets of the application under consumption, is based on deep neural network trained and validated on representative dataset, can be automatically adopted to a new territory.

The approach we are developing is based on a specific architecture of the convolutional network which has had a great success in solving the semantic segmentation problems.
In the mathematical problem statement, a general formal description of the detected changes is introduced for a pair of images; the description being applied to all datasets analyzed. The neural network architecture makes it possible to use the pre-trained parameters of the network and adapt it to new datasets with an additional training at a very little expense. 

The technique can be applied to all pairs of the images in the series; the post-processing is then used to build a more common object that describes the image evolution within the series. The post-processing can be specific for each business application and does not depend on the deep learning technique used at the early stages of analysis.

\section{Conclusion}

In this paper we have summarized the knowledge about the business applications of change detection, known methods of the image processing and deep learning and the possibilities and limitations of the RS data to propose a generalized framework that can embed the solutions for different problems in the area of the targeted change detection in the remote sensing data. The proposed framework to be implemented in our future research will facilitate the integration of the current and future data sources and the advanced algorithms of the data processing.

%

%
%

%
\end{document}